\documentclass{article}
\usepackage[final]{nips_2016}
\usepackage{amsmath,amssymb}
\usepackage{multirow}
\usepackage{graphicx,subfig}
\usepackage{color}
\usepackage{soul}
\newcommand*\samethanks[1][\value{footnote}]{\footnotemark[#1]}

\title{Active Learning for Speech Recognition: \\the Power of Gradients}

\author{
  Jiaji Huang\thanks{Equal contribution.},~~
  Rewon Child\samethanks,~~
  Vinay Rao\samethanks\\
  \texttt{\{huangjiaji, rewon, vinay\}@baidu.com }
  \AND 
  Hairong Liu,~~
  Sanjeev Satheesh,~~
  Adam Coates\\
  \texttt{\{liuhairong, sanjeevsatheesh, adamcoates\}@baidu.com} \\
  Baidu Silicon Valley AI Lab \\
  1195 Bordeaux Dr \\
  Sunnyvale, CA, 94089\\
}

\def\x{{\mathbf x}}
\def\E{{\mathbb E}}
\def\I{{\mathbf I}}
\def\N{{\mathcal N}}
\def\tr{{\rm tr}}

\DeclareMathOperator*{\argmax}{arg\,max}
\def\ben{\begin{equation}}
\def\een{\end{equation}}

\begin{document}
\maketitle
\begin{abstract}
In training speech recognition systems, labeling audio clips can be expensive, and not all data is equally valuable. Active learning aims to label only the most informative samples to reduce cost. For speech recognition, confidence scores and other likelihood-based active learning methods have been shown to be effective. Gradient-based active learning methods, however, are still not well-understood.  This work investigates the Expected Gradient Length (\emph{EGL}) approach in active learning for end-to-end speech recognition. We justify \emph{EGL} from a variance reduction perspective, and observe that \emph{EGL}'s measure of informativeness picks novel samples uncorrelated with confidence scores. Experimentally, we show that \emph{EGL} can reduce word errors by 11\%, or alternatively, reduce the number of samples to label by 50\%, when compared to random sampling.
\end{abstract} 

\section{Introduction}
State-of-the-art automatic speech recognition (ASR) systems~\cite{DS2} have large model capacities and require significant quantities of training data to generalize. 
Labeling thousands of hours of audio, however, is expensive and time-consuming. 
A natural question to ask is how to achieve better generalization with fewer training examples. 
Active learning studies this problem by identifying and labeling only the most informative data, potentially reducing sample complexity.
How much active learning can help in large-scale, end-to-end ASR systems, however, is still an open question.

The speech recognition community has generally identified the informativeness of samples by calculating confidence scores.
In particular, an utterance is considered informative if the most likely prediction has small probability~\cite{Riccardi2005}, or if the predictions are distributed very uniformly over the labels~\cite{max_global_entropy}. 
Though confidence-based measures work well in practice, less attention has been focused on gradient-based methods like Expected Gradient Length (\emph{EGL})~\cite{Settles_survey}, where the informativeness is measured by the norm of the gradient incurred by the instance. \emph{EGL} has previously been justified as intuitively measuring the expected change in a model's parameters~\cite{Settles_survey}.We formalize this intuition from the perspective of asymptotic variance reduction, and experimentally, we show \emph{EGL} to be superior to confidence-based methods on speech recognition tasks.
Additionally, we observe that the ranking of samples scored by \emph{EGL} is not correlated with that of confidence scoring, suggesting \emph{EGL} identifies aspects of an instance that confidence scores cannot capture.

In~\cite{Settles_survey}, \emph{EGL} was applied to active learning on sequence labeling tasks, but our work is the first we know of to apply \emph{EGL} to speech recognition in particular. Gradient-based methods have also found applications outside active learning. For example, \cite{Zhang2015} suggests that in stochastic gradient descent, sampling training instances with probabilities proportional to their gradient lengths can speed up convergence. From the perspective of variance reduction, this importance sampling problem shares many similarities to problems found in active learning.

\section{Problem Formulation}
Denote $\x$ as an utterance and $y$ the corresponding label (transcription).
A speech recognition system models the conditional distribution $p(y|\x,\theta)$, where $\theta$ are the parameters in the model, and $p(y|\x,\theta)$ is typically implemented by a Recurrent Neural Network (RNN).
A training set is a collection of  $(\x,y)$ pairs, denoted as $\{(\x_i,y_i)\}_{i=1}^n$.
The parameters of the model are estimated by minimizing the negative log-likelihood on the training set:
\ben
\hat\theta_n = \min_{\theta}\frac{1}{n}\sum_{i=1}^n \left[\ell(\x_i,y_i,\theta) \triangleq -\log p(y_i|\x_i,\theta)\right].
\een
Active learning seeks to augment the training set with a new set of utterances and labels $\{(\x_i^*, y^*)\}_{i=1}^m$ in order to achieve good generalization on a held-out test dataset. In many applications, there is an unlabeled pool $U$ which is costly to label in its entirety. $U$ is \emph{queried} for the ``most informative'' instance(s) $\x_i^*$, for which the label(s) $y_i^*$ are then obtained.  We discuss several such \emph{query strategies} below.

\subsection{Confidence Scores}
Confidence scoring has been used extensively as a proxy for the informativeness of training samples.
Specifically, an $\x_i^*$ is considered informative if the predictions are uniformly distributed over all the labels~\cite{max_global_entropy}, or if the best prediction of its label is with low probability~\cite{Riccardi2005}. 
By taking the instances which ``confuse'' the model, these methods may effectively explore under-sampled regions of the input space.

\subsection{Expected Gradient Length}
Intuitively, an instance can be considered informative if it results in large changes in model parameters. A natural measure of the change is gradient length, $\|\nabla_\theta\ell(\x_i,y_i;\theta)\|$. 
Motivated by this intuition, Expected Gradient Length (\emph{EGL})~\cite{Settles_survey} picks the instances expected to have the largest gradient length. Since labels are unknown on $U$, \emph{EGL}
computes the expectation of the gradient norm over all possible labelings. 
\cite{Settles_survey} interprets \emph{EGL} as ``expected model change''.
In the following section, we formalize the intuition for \emph{EGL} and show that it follows naturally from reducing the variance of an estimator.

\subsection{Variance in the Asymptote}
Assume the joint distribution of $(\x,y)$ has the following form,
\[
    p(\x,y|\theta_0) = p(y|\x,\theta_0)p(\x),
    \label{eq:jointP}
\]
where $\theta_0$ is the true parameter, and $p(\x)$ is independent of $\theta_0$.
By selecting a subset of the training data, we are essentially choosing another distribution $q(\x)$ so that the $(\x,y)$ pairs are drawn from 
\[
q(\x,y|\theta_0)=p(y|\x,\theta_0)q(\x).
\]
Statistical signal processing theory~\cite{Asympt2016} states the following asymptotic distribution of $\hat\theta_n$,
\ben
\sqrt{n}\left( \hat\theta_n - \theta_0 \right) \rightarrow 
\N(0, \I_q^{-1}(\theta_0)]),
\een
where $\I_q(\theta_0) \mathrel{\stackrel{\mbox{def}}{=}} \E_{q(\x,y)}\left[\nabla_\theta\log p(\x,y|\theta_0)\nabla_\theta^\top\log p(\x,y|\theta_0)\right]$ is the Fisher Information Matrix with respect to $q(\x,y)$. 
Using first order approximation at $\ell(\x,y;\theta_0)$, we have asymptotically, 
\ben
\sqrt{n}(\ell(\x,y;\hat\theta_n)-\ell(\x,y;\theta_0)) \rightarrow \N(0,\nabla_\theta^\top\ell(\x,y;\theta_0)\I_q^{-1}(\theta_0)\nabla_\theta\ell(\x,y;\theta_0)).
\label{eq:asymp}
\een

Eq.~\eqref{eq:asymp} indicates that to reduce $\ell(\x,y;\hat\theta_n)$ on test data, we need to minimize the expected variance $\E_{p(\x,y)}[\nabla_\theta^\top\ell(\x,y;\theta_0)\I_q^{-1}(\theta_0)\nabla_\theta\ell(\x,y;\theta_0)]$ over the test set.
This is called Fisher Information Ratio criteria in \cite{Zhang2000}, which itself is hard to optimize. An easier surrogate is to maximize $\tr(\I_q(\theta_0))$. 
Substituting Eq.~\eqref{eq:jointP} into $\I_q(\theta_0)$, we have
\[
    \I_q(\theta_0) = \E_{q(\x,y)}\left[\nabla_\theta\log p(y|\x,\theta_0) \nabla_\theta^\top\log p(y|\x,\theta_0)\right] = \E_{q(\x,y)}\left[ \nabla_\theta \ell(\x,y;\theta_0) \nabla_\theta^\top \ell(\x,y;\theta_0)\right],
\]
which is equivalent to
$
\max_{q} \int q(\x) \int p(y|\x,\theta_0)\|\nabla_{\theta} \ell(\x,y;\theta_0)\|^2 dy d\x.
$

A practical issue is that we do not know $\theta_0$ in advance. We could instead substitute an estimate $\hat \theta_0$ from a pre-trained model, where it is reasonable to assume the $\hat\theta_0$ to be close to the true $\theta_0$.
The batch selection then works by taking the samples that have largest gradient norms,
\ben
i^\ast = \argmax_i \sum_y p(y|\x_i,\hat\theta_0) \|\nabla_{\theta} \ell(\x_i,y;\hat\theta_0)\|^2.
\label{eq:maxGNorm}
\een
For RNNs, the gradients for each potential label can be obtained by back-propagation. Another practical issue is that \emph{EGL} marginalizes over all possible labelings, but in speech recognition, the number of labelings scales exponentially in the number of timesteps. Therefore, we only marginalize over the $K$ most probable labelings. They are obtained by beam search decoding, as in~\cite{Settles2008}. The \emph{EGL} method in \cite{Settles_survey} is almost the same as Eq.~\eqref{eq:maxGNorm}, except the gradient's norm is not squared in~\cite{Settles_survey}. 

Here we have provided a more formal characterization of \emph{EGL} to complement its intuitive interpretation as ``expected model change'' in~\cite{Settles_survey}. For notational convenience, we denote Eq.~\eqref{eq:maxGNorm} as \emph{EGL} in subsequent sections.

\section{Experiments}
We empirically validate \emph{EGL} on speech recognition tasks. In our experiments, the RNN takes in spectrograms of utterances, passing them through two 2D-convolutional layers, followed by seven bi-directional recurrent layers and a fully-connected layer with softmax activation. All recurrent layers are batch normalized. 
At each timestep, the softmax activations give a probability distribution over the characters.
CTC loss~\cite{GravesCTC} is then computed from the timestep-wise probabilities.

A base model, $\hat\theta_0$, is trained on 190 hours ($\sim$100K instances) of transcribed speech data. Then, it selects a subset of a 1,700-hour ($\sim$1.1M instances) unlabeled dataset. We query labels for the selected subset and incorporate them into training. Learning rates are tuned on a small validation set of 2048 instances. The trained model is then tested on a 156-hour ($\sim$100K instances) test set and we report CTC loss, Character Error Rate (CER) and Word Error Rate (WER).

The confidence score methods~\cite{Riccardi2005, max_global_entropy} can be easily extended to our setup. Specifically, from the probabilities over the characters, we can compute an entropy per timestep and then average them. This method is denoted as \emph{entropy}. 
We could also take the most likely prediction and calculate its CTC loss, normalized by number of timesteps. This method is denoted as \emph{pCTC} (predicted CTC) in the following sections.
\begin{figure}[h!]
\centering
\subfloat[CTC loss on dev set]{\includegraphics[width=0.33\textwidth]{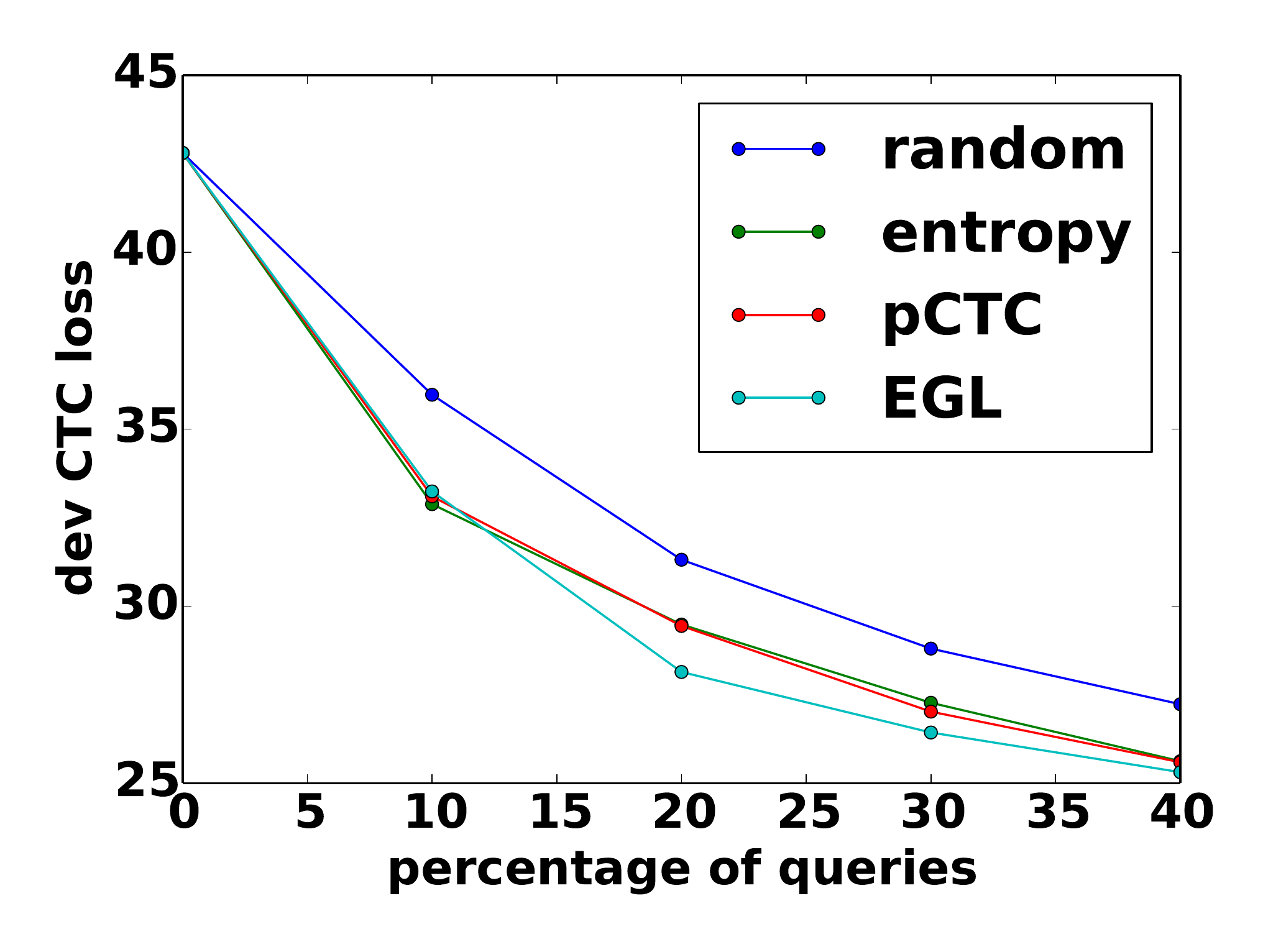}}
\subfloat[CER on dev set]{\includegraphics[width=0.33\textwidth]{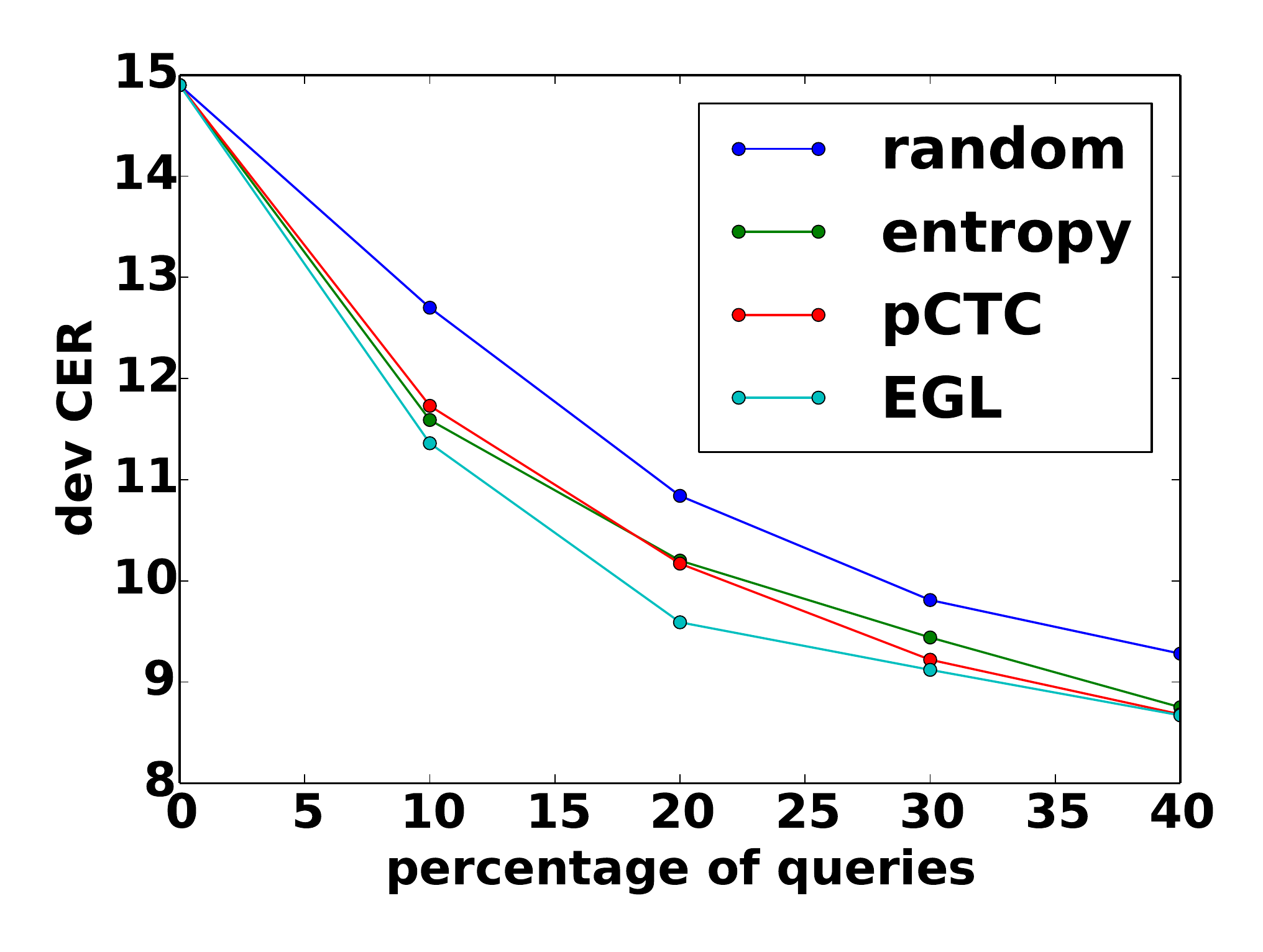}}
\subfloat[WER on dev set]{\includegraphics[width=0.33\textwidth]{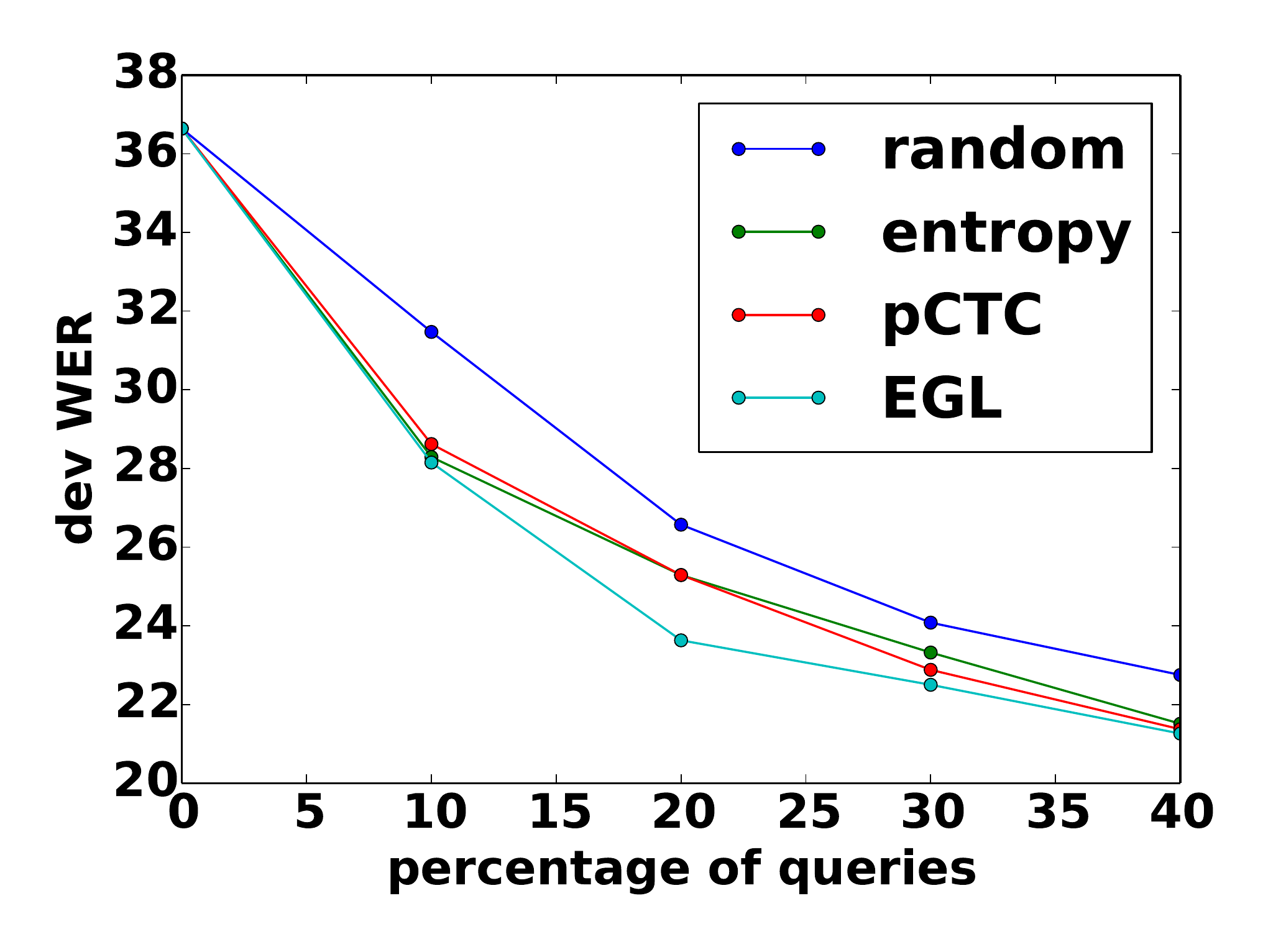}}
\caption{\small Performance metrics at various percentages of queries. \emph{EGL} shows a greater reduction in error for smaller amounts of data. By definition, all strategies converge as the query percentage approaches 100\%.}
\label{fig:metrics}
\end{figure}

We implement \emph{EGL} by marginalizing over the most likely 100 labels, and compare it with: 1) a \emph{random} selection baseline,
2) \emph{entropy}, and 3) \emph{pCTC}. Using the same base model, each method queries a variable percentage of the unlabeled dataset.
The queries are then included into training set, and the model continues training until convergence.
Fig.~\ref{fig:metrics} reports the metrics (Exact values are reported in Table~\ref{tab:metrics} in the Appendix) on the test set as the query percentage varies.
All the active learning methods outperform the \emph{random} baseline. 
Moreover, \emph{EGL} shows a steeper, more rapid reduction in error than all other approaches. Specifically, when querying 20\% of the unlabeled dataset, \emph{EGL} has 11.58\% lower CER and 11.09\% lower WER relative to \emph{random}. The performance of \emph{EGL} at querying 20\% is on par with \emph{random} at 40\%, suggesting that using \emph{EGL} can lead to an approximate 50\% decrease in data labeling.

\subsection{Similarity between Query Methods}
It is useful to understand how the three active learning methods differ in measuring the informativeness of an instance. To compare any two methods, we take rankings of informativeness given by these two methods, and plot them in a 2-D ranking-vs-ranking coordinate system. A plot close to the diagonal implies that these two methods evaluate informativeness in a very similar way.

\begin{figure}[h!]
    \centering
    \subfloat[\emph{pCTC} rankings vs \emph{entropy} rankings]{\label{fig:pctc-ent}\includegraphics[width=0.35\textwidth]{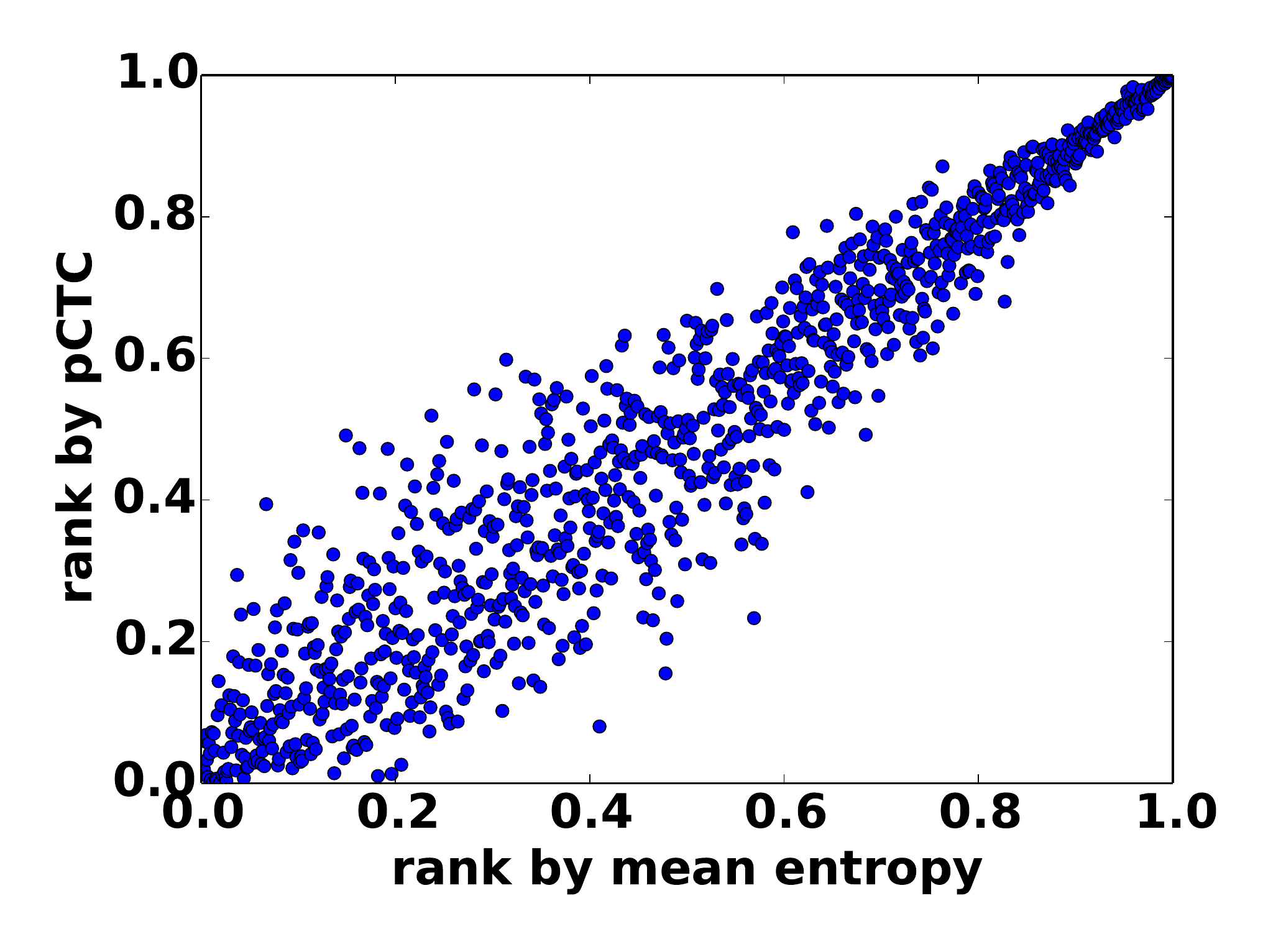}}
    \qquad\qquad
    \subfloat[\emph{EGL} rankings vs \emph{entropy} rankings]{\label{fig:egl-ent}\includegraphics[width=0.35\textwidth]{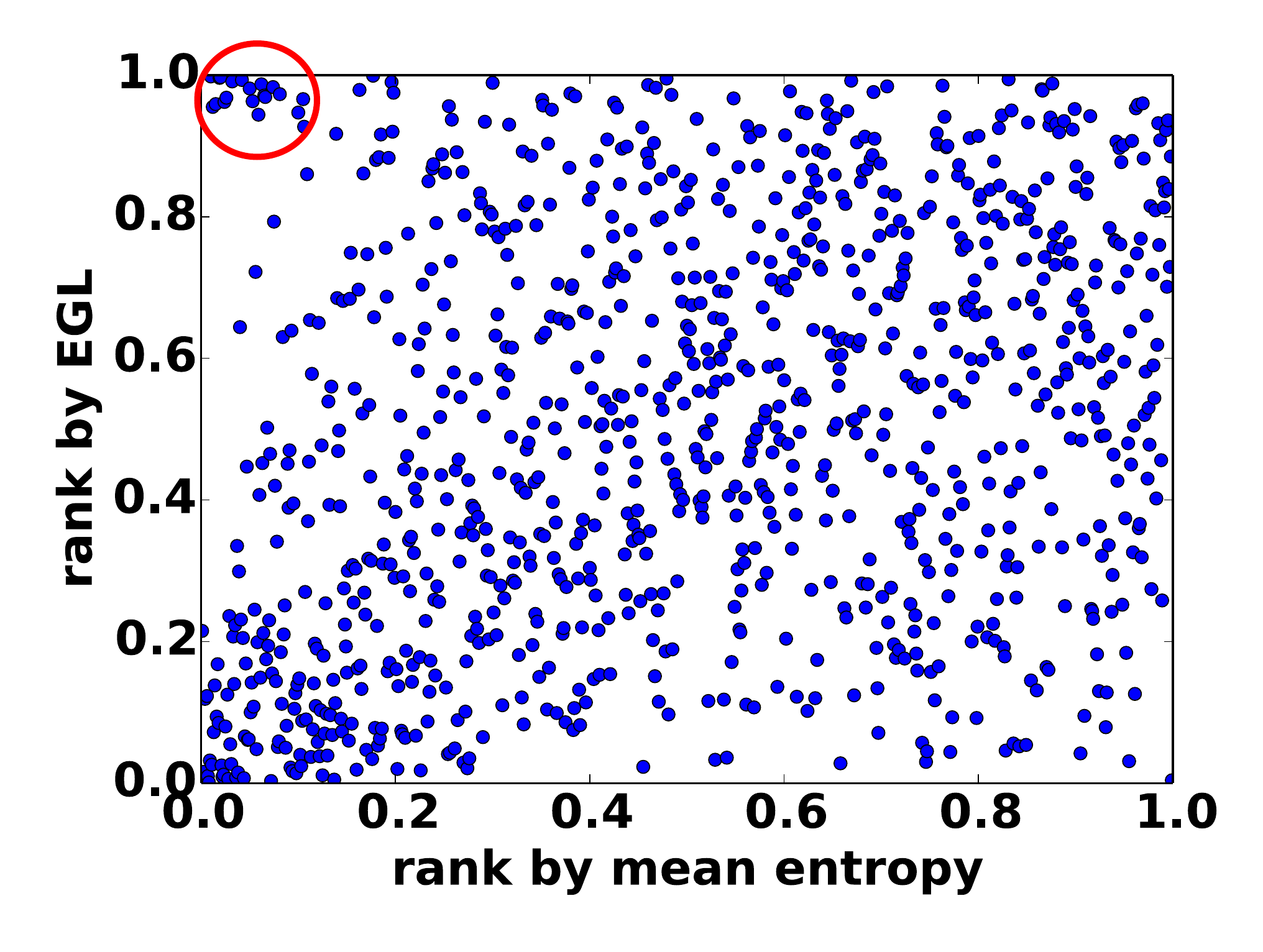}}
    \caption{\small The difference in how active learning methods rank informativeness of samples. Rankings are normalized to $[0,1]$, with 1 being the most informative. In (a), \emph{pCTC} and \emph{entropy} are shown to be very correlated. In (b), \emph{EGL} appears uncorrelated with \emph{entropy} (and \emph{pCTC}). Data samples highlighted in the red circle are considered very informative by \emph{EGL}, but uninformative by \emph{entropy}.}
    \label{fig:rank_correlations}
\end{figure}

Fig.~\ref{fig:rank_correlations} shows the ranking-vs-ranking plots between \emph{pCTC} and \emph{entropy}, \emph{EGL} and \emph{entropy}. We observe that \emph{pCTC} rankings and \emph{entropy} rankings (Fig.~\ref{fig:pctc-ent}) are very correlated. This is likely because they are both related to model uncertainty. In contrast, \emph{EGL} gives very different rankings from \emph{entropy} (Fig.\ref{fig:egl-ent}). This suggests \emph{EGL} is able to identify aspects of an instance that uncertainty-based measurements cannot capture.

We further investigate the samples for which \emph{EGL} and \emph{entropy} yield vastly different estimates of informativeness, e.g., the elements in the red circle in Fig.~\ref{fig:egl-ent}. These particular samples consist of short utterances containing silence (with background noise) or filler words. Further investigation is required to understand whether these samples are noisy outliers or whether they are in fact important for training end-to-end speech recognition systems.

\section{Conclusion and Future Work}
We formally explained \emph{EGL} from a variance reduction perspective and experimentally tested its performance on end-to-end speech recognition systems.
Initial experiments show a notable gain over random selection, and that it outperforms confidence score methods used in the ASR community. We also show \emph{EGL} measures sample informativeness in a very different way from confidence scores, giving rise to open research questions.
All the experiments reported here query all samples in a single batch. It is also worth considering the effects of querying samples in a sequential manner. In the future, we will further validate the approach with sequential queries and seek to make the informativeness measure robust to outliers.

\section*{Appendix}
\setlength{\tabcolsep}{2.3pt}
\begin{table}[h!]
    \centering
    \caption{Performance metrics at various query percentages (smaller is better, best in bold)}
    \begin{tabular}{c||c|c|c|c||c|c|c|c||c|c|c|c}
        \hline\hline
         \multirow{ 2}{*}{query} & \multicolumn{4}{c||}{CTC} & \multicolumn{4}{c||}{CER} & \multicolumn{4}{c}{WER} \\
         \cline{2-13}
         & \emph{\small random} & \emph{\small entropy} & \emph{\small pCTC} & \emph{\small EGL} & \emph{\small random} & \emph{\small entropy} & \emph{\small pCTC} & \emph{\small EGL} & \emph{\small random} & \emph{\small entropy} & \emph{\small pCTC} & \emph{\small EGL}\\
         \hline 
         10\% & 35.97 & {\bf 32.88} & 33.10 & 33.24 & 12.70 & 11.59 & 11.73 & {\bf 11.36} & 31.47 & 28.29 & 28.62 & {\bf 28.15} \\
         \hline
         20\% & 31.31 & 29.48 & 29.44 & {\bf 28.14} & 10.84 & 10.20 & 10.17 & {\bf 9.59} & 26.57 & 25.29 & 25.29 & {\bf 23.63} \\
         \hline
         30\% & 28.80 & 27.27 & 27.02 & {\bf 26.43} & 9.81 & 9.44 & 9.22 & {\bf 9.12} & 24.08 & 23.32 & 22.88 & {\bf 22.50} \\
         \hline
         40\% & 27.23 & 25.62 & 25.59 & {\bf 25.31} & 9.28 & 8.75 & 8.68 & {\bf 8.67} & 22.75 & 21.51 & 21.37 & {\bf 21.26} \\
         \hline\hline 
    \end{tabular}
    \label{tab:metrics}
\end{table}

\bibliographystyle{plain}
\bibliography{refs}
\end{document}